\documentclass{ecai} 

\usepackage{latexsym}
\usepackage{amssymb}
\usepackage{amsmath}
\usepackage{amsthm}
\usepackage{booktabs}
\usepackage{enumitem}
\usepackage{graphicx}
\usepackage{color}
\usepackage{cite}
\usepackage{algorithm}
\usepackage{algpseudocode}
\usepackage{textcomp}
\usepackage{color, colortbl}
\usepackage{multirow}
\usepackage{mdframed}
\usepackage{xcolor}
\usepackage{url}
\usepackage{booktabs}

\newcommand{\BibTeX}{B\kern-.05em{\sc i\kern-.025em b}\kern-.08em\TeX}

\begin{document}


\begin{frontmatter}


\paperid{4605} 


\title{GeoSAM: Fine-tuning SAM with Multi-Modal \\
Prompts for Mobility Infrastructure Segmentation}


\author[A]{\fnms{Rafi Ibn}~\snm{Sultan}}
\author[B]{\fnms{Chengyin}~\snm{Li}}
\author[A]{\fnms{Hui}~\snm{Zhu}}
\author[A]{\fnms{Prashant}~\snm{Khanduri}}
\author[C]{\fnms{Marco}~\snm{Brocanelli}}
\author[A]{\fnms{Dongxiao}~\snm{Zhu}\thanks{Corresponding Author. Email: dzhu@wayne.edu.}}

\address[A]{Department of Computer Science, Wayne State University, Detroit, MI, USA 48202} 
\address[B]{Department of Radiation Oncology, Henry Ford Health, Detroit, MI, USA 48202} 
\address[C]{Department of Electrical and Computer Engineering, The Ohio State University, Columbus, Ohio, USA 43210}


\begin{abstract}
In geographical image segmentation, performance is often constrained by the limited availability of training data and a lack of generalizability, particularly for segmenting mobility infrastructure such as roads, sidewalks, and crosswalks. Vision foundation models like the Segment Anything Model (SAM), pre-trained on millions of natural images, have demonstrated impressive zero-shot segmentation performance, providing a potential solution. However, SAM struggles with geographical images, such as aerial and satellite imagery, due to its training being confined to natural images and the narrow features and textures of these objects blending into their surroundings. To address these challenges, we propose Geographical SAM (GeoSAM), a SAM-based framework that fine-tunes SAM using automatically generated multi-modal prompts. Specifically, GeoSAM integrates point prompts from a pre-trained task-specific model as primary visual guidance, and text prompts generated by a large language model as secondary semantic guidance, enabling the model to better capture both spatial structure and contextual meaning. GeoSAM outperforms existing approaches for mobility infrastructure segmentation in both familiar and completely unseen regions by at least 5\% in mIoU, representing a significant leap in leveraging foundation models to segment mobility infrastructure, including both road and pedestrian infrastructure in geographical images. The source code is publicly available$^{1}$. 

\end{abstract}

\end{frontmatter}


\section{Introduction}
\label{sec:intro}\footnote{Code available at \url{https://github.com/rafiibnsultan/GeoSAM}} 
While a substantial amount of research~\citep{CHEN2022103004,8575492, Henry_2018, 10.1007/978-3-031-18458-1_51, Gudžius2021} has focused on road infrastructure segmentation from geographical and remote sensing imagery like aerial and satellite images, pedestrian infrastructure, such as sidewalks or crosswalks, has received comparatively little attention, despite its importance in daily life. Historically, research efforts have predominantly focused on assisting drivers in navigation rather than pedestrians~\citep{hosseini2023mapping}. Existing accessibility studies often use simplified road data, but accurate segmentation of pedestrian infrastructure can better reveal accessible routes and destinations, especially for people with disabilities.

Rooted in historical context, mobility infrastructure segmentation has predominantly relied on traditional models, including Convolutional Neural Networks (CNNs)~\citep{sun2019deep, zhou2018unet++, isensee2021nnu, ronneberger2015unet} and Vision Transformer (ViT) models~\citep{hatamizadeh2021swin, dosovitskiy2021vit}. These models typically require large collections of human-labeled data for task-specific training~\citep{hosseini2023mapping, cha2023billionscale}, something that is oftentimes a luxury for these tasks, and are often too sensitive to changes in data. However, the scarcity of high-quality labeled datasets remains a major challenge, especially in the context of mobility infrastructure, limiting scalability and adaptability to diverse tasks.

\begin{figure}[t]
    \centering
    \includegraphics[width=0.9\linewidth]{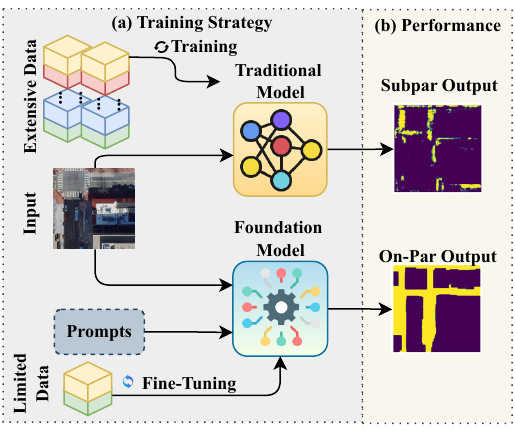}
    \caption{Mobility infrastructure segmentation: (a) Traditional models need large task-specific datasets, (b) struggle with narrow, texture-similar objects, yielding subpar results. Fine-tuning a promptable foundation model with limited data and prompts achieves on-par performance.}
    \label{fig:idea_figure}
\end{figure}

Traditional models, when trained on limited and homogeneous datasets(Figure~\ref{fig:idea_figure}a), often fail to distinguish fine-grained classes such as sidewalks and roads, which exhibit subtle visual differences like thin boundaries, similar textures, and frequent occlusions (Figure~\ref{fig:idea_figure}b). Moreover, their learned representations are typically domain-specific, resulting in poor generalization when deployed in unseen regions or datasets with different visual characteristics. Even minor shifts in data distribution, such as moving from one geographic region to another, often lead to significant performance degradation. In contrast, vision foundation models, pre-trained task-agnostically on large-scale and diverse image distributions~\citep{kirillov2023segment, radford2021learning}, offer a promising alternative with superior generalization ability across varying domains. These models adapt to new downstream tasks without re-training, relying on user-provided prompts for contextual guidance. In this work, we leverage the Segment Anything Model (SAM)~\citep{kirillov2023segment}, a promptable vision foundation model, to overcome the limitations of traditional approaches and enable effective segmentation of mobility infrastructure, even with limited labeled data and across geographically diverse regions.





However, unlike compact objects in natural images, where a single point (e.g., placed on a dog’s body) often suffices for segmentation, spatially extensive structures like roads and sidewalks usually require multiple iterative prompts to capture their full extent. This process is often exhaustive and error-prone, and even with multiple prompts, zero-shot SAM struggles in remote sensing tasks due to its pre-training on natural images, which lack the large, texture-similar structures common in geographical data~\citep{kirillov2023segment}. Nonetheless, SAM’s general segmentation capability can be adapted to geographical imagery via fine-tuning on limited data (bottom of Figure~\ref{fig:idea_figure}), allowing it to learn domain-specific patterns and remain effective under regional distribution shifts. Capitalizing on this strength, we introduce Geographical SAM (GeoSAM), an end-to-end model tailored for segmenting mobility infrastructure through multi-class segmentation of road and pedestrian infrastructure. 

To address these challenges, we propose Geo-Point Generation (GPG), an automated prompt generation technique that generates point prompts for geographical images from a domain-specific pre-trained model for precise spatial guidance. It is complemented by text prompts for semantic clarity to resolve ambiguities inherent in point-based guidance. Point prompts focus the model on specific pixels, but a single pixel can often belong to multiple objects. Text prompts, containing semantic information about the class, clarify the object of interest and provide a broader understanding~\citep{yan2024urbanclip}. This complementary design ensures precise geometrical guidance from point prompts and broader contextual understanding from text prompts, enhancing segmentation accuracy.


These multi-modal prompts fine-tune SAM through its lightweight decoder. By integrating spatial precision with semantic context, we introduce \textbf{Geographical SAM (GeoSAM)}, an end-to-end SAM-based model fine-tuned for multi-class segmentation of roads and pedestrian infrastructure. GeoSAM outperforms traditional CNN-based approaches~\citep{hosseini2023mapping, Henry_2018, 8575492}, not only improving segmentation accuracy but also demonstrating the potential of combining natural language and visual interaction within foundation models for geographical imagery. Our contributions are three-fold: (1) We pioneer the use of SAM for multi-class mobility infrastructure segmentation, integrating point and text prompts in geographical imagery. (2) We introduce fine-tuning and automated prompt generation techniques that inject domain knowledge from traditional models via multi-modal prompts. (3) We conduct extensive evaluations on datasets from two cities, demonstrating GeoSAM's strong performance and generalizability across diverse locations.

\begin{figure*}[t]
\centering
\includegraphics[width=1.0\textwidth]{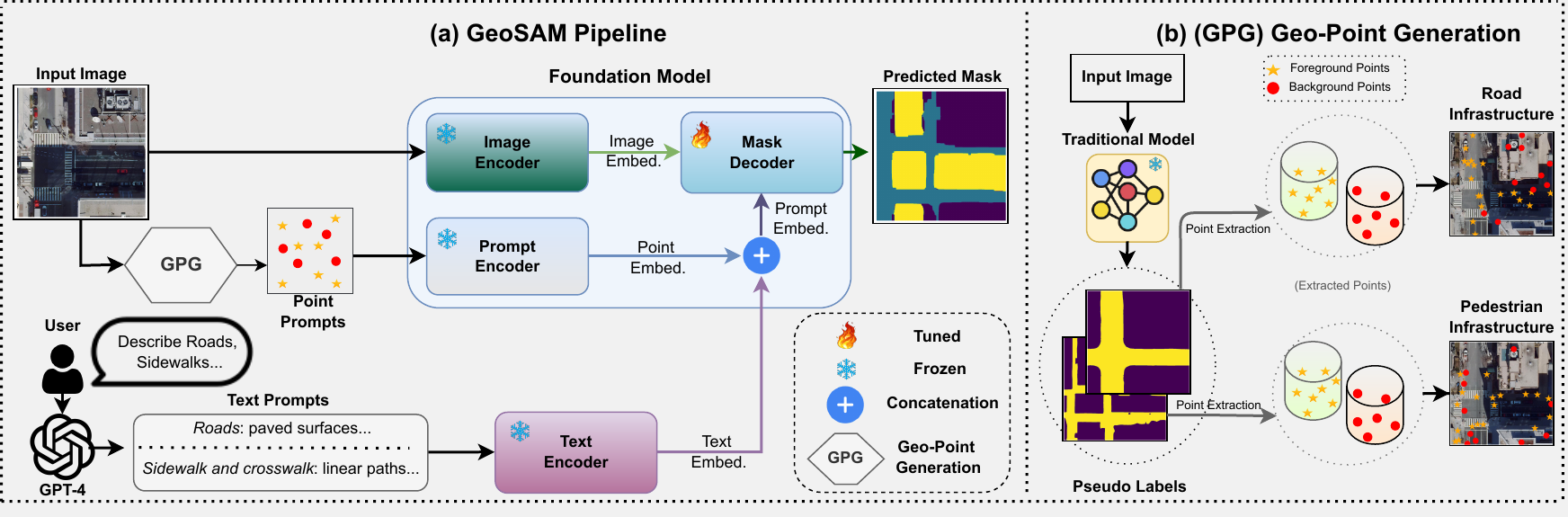}
\vspace{-10pt}
\caption{(a) \textbf{GeoSAM}: Training pipeline for mobility infrastructure segmentation.  Text prompts generated by GPT-4 and point prompts generated from a pre-trained traditional model are utilized to tune a foundation model decoder in producing segmentation masks (yellow = road, blue = pedestrian). (b) \textbf{Geo-Point Generation (GPG)}: Point prompts are generated from pseudo labels created by a pre-trained traditional model; \textcolor{orange}{stars} and \textcolor{red}{circles} represent foreground and background points, respectively.}
\label{fig:GeoSAM}

\end{figure*}

\section{Related Work}

\label{sec:related}
\subsection{Traditional Geographical Methods}
Before the emergence of foundation models, traditional task-specific works, such as UNet-based approaches like~\citep{Henry_2018, 10.1007/978-3-031-18458-1_51} and more advanced encoder-decoder-based works like~\citep{8575492, CHEN2022103004, Gudžius2021} were developed to execute various geographical image segmentation tasks. Furthermore, CNN-based work such as~\citep{hosseini2023mapping} focuses more on pedestrian infrastructure segmentation in aerial images. Researchers have also explored machine learning techniques to enhance CNN-based segmentation for geographic objects~\citep{7913730, ayush2021geography}, along with transfer learning approaches that leverage pretrained models~\citep{8100027}. While these efforts improve remote sensing segmentation, they often rely on extensive supervision and retraining. Accuracy gains aside, they fall short in addressing the core challenge of generalizing to new locations.


 


\subsection{Geographical Foundation Models} 

Task-agnostic vision foundation models address traditional segmentation limitations by using prompts to adapt to unseen classes across diverse tasks. While their use in geographical imagery, such as SAM, remains limited, some studies have begun exploring its potential. Works like~\citep{wang2023scalingup, ma2024sam, osco2023segment} leverage SAM's zero-shot capabilities for tasks beyond segmentation, with~\citep{osco2023segment} employing a hybrid zero-shot and one-shot learning approach for geographical imagery segmentation. However, these approaches are largely effective for objects with well-defined boundaries and distinguishable physical contexts, relying primarily on sensible prompts without requiring extensive domain-specific knowledge.

Most research focuses on manual human prompting during inference, though automated prompt generation has gained attention. Studies like~\citep{chen2024rsprompter, wang2023cs, zhang2024uv} develop automated prompt-generation techniques requiring substantial training data, while others, such as~\citep{zhang2023text2seg, osco2023segment}, use text queries in two-stage pipelines to generate bounding box prompts for SAM. Direct integration of natural language text prompts for improving SAM in geographical imagery remains unexplored. Methods like \citep{liu2024rsps, zhou2024mesam} eliminate the need for prompts using additional networks, but require extensive training data, while \citep{wu2025tpp} depends on auxiliary inputs like trajectory points, making it road-specific and less generalizable to other classes.

To address domain-specific challenges, some works have fine-tuned SAM using Parameter Efficient Fine-Tuning (PEFT) techniques~\citep{hu2021lora}. In geographical imagery, studies like~\citep{yan2023ringmo, ding2023adapting, chen2024rsprompter, feng2024road} explore fine-tuning for diverse downstream tasks, yet no work, to our knowledge, focuses on fine-tuning SAM specifically for mobility tasks such as pedestrian infrastructure segmentation. This critical gap presents an opportunity for significant social impact in underperforming tasks.



\subsection{Domain-Specific Geographical Foundation Models} 
Researchers have also explored training domain-specific foundation models on large-scale geographical imagery for targeted tasks. Similar to SAM, works like~\citep{cha2023billionscale, sun2022ringmo} develop non-promotable foundation models using scaled ViT architectures, focusing on specific tasks without user interaction. In a related effort,~\citep{yan2023ringmo} employs a SAM-like architecture trained on a massive remote sensing dataset. While many of these studies target road segmentation, they overlook the critical task of mobility infrastructure segmentation, such as sidewalks and crosswalks. Moreover, the lack of public source code makes it difficult to evaluate their effectiveness for pedestrian infrastructure.


\section{Method}
\label{sec:method}

\subsection{Problem Definition}
Given a geographical or remote sensing image i.e. aerial or satellite imagery dataset \(D\) containing \(n\) sample images, where each image  \(I \in \mathbb{R}^{H \times W \times 3}\) represents a standard high-resolution RGB image with height \(H\), width \(W\), and 3 color channels. We implement GeoSAM (illustrated in Figure~\ref{fig:GeoSAM}a), which produces a multi-class segmentation map \( S \in \mathbb{R}^{H \times W} \) for each input image, where each pixel stores the predicted class index (e.g., background, pedestrian infrastructure, or road infrastructure). For training purposes, we convert this into a one-hot encoded multi-channel representation \( \hat{S} \in \{0, 1\}^{H \times W \times C} \), where \( C \) is the number of classes, and \( \hat{S}_{i,j,c} = 1 \) if pixel \( (i,j) \) belongs to class \( c \), and 0 otherwise.

\subsection{SAM: Background}
Segment Anything Model (SAM) consists of an image encoder (\( \operatorname{Enc_I} \)), a prompt encoder (\( \operatorname{Enc_P} \)), and a mask decoder (\( \operatorname{Dec_M} \)). Given an input image \( I \in \mathbb{R}^{H \times W \times 3} \) and a set of prompts \( P \), SAM encodes the image as \( F_I = \operatorname{Enc_I}(I) \) and the prompts as \( T_P = \operatorname{Enc_P}(P) \). These embeddings are passed to the decoder, which performs attention-based interactions and predicts the segmentation mask:
\[
S = \operatorname{Dec_M}(F_I, T_P).
\]
Prompt embeddings \( T_P \) guide the decoder using both spatial and semantic information.



\subsection{Multi-Modal Prompt Generation}



\noindent\textbf{Point Prompts} To segment sparse and spatially extensive structures like roads and sidewalks in remote sensing images, GeoSAM leverages point prompts as its primary guidance mechanism. Unlike bounding boxes, which are often impractical for such objects due to their large spatial extent, point prompts enable precise localization with minimal input. We introduce GPG, an automated approach (Figure~\ref{fig:GeoSAM}b) that generates multiple foreground and background point prompts from a pre-trained traditional model \( f_{\text{pre}} \). Each point is a 2D spatial coordinate on the image, serving as either a foreground cue to guide the model’s attention or a background cue to indicate regions to avoid. Using multiple points helps reduce ambiguity and enables the model to more accurately localize the target object, especially in complex or overlapping areas.


The input image \( I \in \mathbb{R}^{H \times W \times 3} \) is processed by the pre-trained model \( f_{\text{pre}} \), which outputs a pseudo-label segmentation map of the same size:

\begin{equation}
M_{\text{pseudo}} = f_{\text{pre}}(I), \quad M_{\text{pseudo}}[i, j] \in \{0, 1, \dots, C_{\text{pre}} - 1\},
\end{equation}

\noindent where \( C_{\text{pre}} \) is the number of semantic classes in the pre-trained model, and each pixel in \( M_{\text{pseudo}} \) is assigned one of these class labels. 

We decompose the pseudo-label map \( M_{\text{pseudo}} \) into multiple binary masks, one for each semantic class, as illustrated in Figure~\ref{fig:GeoSAM}b. For each class-specific mask, we randomly sample a set of point prompts \( x = \{ x_i \}_{i=1}^{k} \), where each point \( x_i \) is a 2D coordinate in the image domain \( \Omega_I \subset \mathbb{R}^2 \). The total number of sampled points per class is denoted by \( k \), and these points are used as point prompts to guide the segmentation model. Then the set \( x \) is partitioned into foreground and background points:

\begin{equation}
\begin{aligned}
x^{\text{fg}} &= \{ x_i \in \Omega_I \mid M_{\text{pseudo}}(x_i) \in \mathcal{C}_{\text{fg}} \}, &\quad |x^{\text{fg}}| = k_1, \\
x^{\text{bg}} &= \{ x_i \in \Omega_I \mid M_{\text{pseudo}}(x_i) \in \mathcal{C}_{\text{bg}} \}, &\quad |x^{\text{bg}}| = k_2, \\
x &= x^{\text{fg}} \cup x^{\text{bg}}, 
\end{aligned}
\end{equation}

\noindent where \( k = k_1 + k_2 \) denotes the total number of sampled point prompts and \( \mathcal{C}_{\text{fg}} \) and \( \mathcal{C}_{\text{bg}} \) are the sets of class labels corresponding to foreground and background, respectively. These point prompts are transformed into high-dimensional embeddings of dimension \( C \) using the frozen prompt encoder \( \operatorname{Enc_P} \).

\begin{equation}
T_{{x}} = \operatorname{Enc_P}(x) \in \mathbb{R}^{k \times C}.
\end{equation}

\noindent The pre-learned position embeddings from SAM's pre-training (indicating whether a point is in the foreground or background) are appended with the $T_{x}$. Here, the accuracy of  \( M_{\text{pseudo}} \) is not critical; as long as the generated points are approximately within the foreground or background regions of the class, they can effectively guide the focus of the segmentation process. For our experiments, we adopt a standard semantic segmentation model \( f_{\text{pre}} \) as the pre-trained traditional model, which produces a multi-class segmentation map containing class sets of:
\noindent $
\mathcal{C}_{\text{fg}}^{\text{road}} = \{\text{road}\}, \quad \mathcal{C}_{\text{fg}}^{\text{pedestrian}} = \{\text{sidewalk}, \text{crosswalk}\}, \quad \mathcal{C}_{\text{bg}} = \{\text{background}\}.
$
\noindent Each binary segmentation task, either between \( \mathcal{C}_{\text{fg}}^{\text{road}} \) and \( \mathcal{C}_{\text{bg}} \), or between \( \mathcal{C}_{\text{fg}}^{\text{pedestrian}} \) and \( \mathcal{C}_{\text{bg}} \), uses its corresponding foreground and background point prompts to provide GeoSAM with class-specific spatial guidance.

\noindent\textbf{Text Prompts} In addition to geometric guidance from point prompts, GeoSAM incorporates semantic context through text prompts \(t\), enhancing the model's ability to distinguish between overlapping objects. While point prompts indicate specific pixel locations, a single pixel may belong to multiple classes (e.g., both road and crosswalk at the same time). To resolve such ambiguities, text prompts provide class-specific descriptions of the target object. 

Each text prompt follows the format: ``[class]: Description.'', where [class] denotes the target class (e.g., roads or sidewalks/crosswalks). For example, a generated prompt might be ``Roads: paved surfaces, vehicle lanes'' (as can be seen in Figure~\ref{fig:GeoSAM}a). To improve robustness and avoid overfitting to static descriptions, we dynamically generate diverse class-specific text prompts during training using OpenAI’s GPT-4~\citep{openai2023gpt}. These prompts are generated solely based on the class name, without any access to image content, ensuring variability~\citep{derakhshani2023variational} in phrasing while maintaining class relevance. The following instruction is provided to GPT-4 to create these class definitions:


\begin{mdframed}[backgroundcolor=lightgray]

``role: system, content: You are a creative assistant, skilled in providing detailed visual descriptions of objects as seen in aerial imagery.''

\noindent ``role: user, content: Print out a visual description (don't mention their names) that can be seen from aerial images of [CLS] (in one line, 4 to 5 words, not more, not less).''
\end{mdframed}

For each class, we generate a set of \( t_n \) text prompts \( t = \{ t_1, t_2, \dots, t_{t_n} \} \), which are encoded using CLIP’s text encoder~\citep{radford2021learning} to produce text embeddings \( T_t \in \mathbb{R}^{t_n \times C} \). CLIP (Contrastive Language–Image Pretraining) is a vision-language model trained to align image and text pairs in a shared embedding space. We leverage CLIP’s inherent ability to project text and image inputs into a shared embedding space for effective cross-modal alignment. To enhance class discrimination, we append a learnable class-specific embedding \( E_{\text{cls}} \in \mathbb{R}^{1 \times C} \) to each text embedding, where \( C = 512 \) is the embedding dimension of CLIP. This mitigates the variability introduced by natural language descriptions (e.g., from ChatGPT) by allowing the model to learn a consistent, discriminative representation for each class. The resulting embeddings are L2-normalized along the feature dimension and subsequently projected to match SAM’s embedding dimension of 256 using a trainable linear projection layer \( W_t \in \mathbb{R}^{C \times 256} \):




\begin{equation}
T_t = \text{{NORM}}[f_{\text{clip}}(t)] W_t \in \mathbb{R}^{n \times 256}.
\end{equation}

\noindent\textbf{Joint Multi-Modal Prompts} 
As text prompt embeddings \( T_{{t}} \) encode the semantic representation of the target class, they naturally complement the foreground point embeddings \( T_{{x}} \), which capture precise spatial localization. The two types of prompt embeddings are then concatenated along the batch (prompt) dimension to form the joint prompt embedding: \[
T_P = \begin{bmatrix} T_x \\ T_t \end{bmatrix} \in \mathbb{R}^{(k + n) \times C}.
\]
This design allows \( T_{{x}} \) to provide geometric position cues, while \( T_{{t}} \) enriches the representation with high-level semantic information, enabling the model to reason more effectively about the target object. Then, $T_{P}$ along with $F_I$ are concatenated and supplied to the decoder. 


\subsection{Fine-Tuning the Decoder} 
\label{fine-tuning}
\noindent \textbf{Decoder Architecture} The decoder utilizes a combination of bidirectional transformers, where image embeddings ($F_I$) are updated through repeated Self Attention (SA) and Cross Attention (CA) with prompts. The self-attention operation on \(T_P\) enables interaction among different prompts, allowing them to exchange information and refine their representations before attending to the image features.

\begin{equation}
\begin{aligned}
T'_{P}&=\text{\textbf{SA}}(T_{P}),\\
\hat{T_{P}} &= T'_{P}+\text{\textbf{MLP}}_P(\text{\textbf{CA}}(T'_{P}, F_I)),\\
\hat{F_I} &= F_I+\text{\textbf{MLP}}_I(\text{\textbf{CA}}(F_I, \hat{T_P})), 
\end{aligned}
\end{equation}

\noindent where $\hat{T_P}$ represents the refined set of prompt embeddings, and $\hat{F_I}$ denotes the updated set of visual embeddings after refining the embeddings by attending to the positional and semantic information of the prompts, enabling context-aware representations that ultimately produce the segmentation map. GeoSAM fine-tunes only the decoder while keeping the rest of the model frozen; a common strategy in foundation model adaptation~\citep{hu2021lora}. This PEFT approach leverages the encoder’s general representation capabilities while reducing the computational overhead of a large foundation model by restricting updates to the task-specific decoder.


\noindent \textbf{Segmentation Map Adaptation} SAM is originally a binary-class segmentation model, producing a map that distinguishes only foreground from background for a single class. GeoSAM extends this by generating a multi-channel segmentation map, where each channel corresponds to a target class, such as road or pedestrian infrastructure. Both classes are processed jointly throughout the pipeline by a single shared decoder, and the model outputs all channels simultaneously. This design makes the framework easily extensible to additional classes, and the loss is computed by comparing the resulting multi-channel outputs with one-hot encoded ground truth maps.

\noindent \textbf{Loss Function} We employ Dice Focal Loss, a synergistic combination of Dice Loss and Focal Loss, to address the challenges inherent in segmenting high-resolution remote sensing images. Given that mobility infrastructure occupies only a small fraction of these images, Dice Focal Loss effectively balances the need for precise overlap accuracy while mitigating the impact of severe class imbalance.

\noindent Let \( S^c \in [0, 1]^N \) and \( G^c \in \{0, 1\}^N \) denote the predicted and ground-truth binary masks for class \( c \), flattened over all \( N \) pixels. Dice Loss is defined as:
\begin{equation}
\mathcal{L}_{\text{Dice}} = 1 - \frac{2 \sum_{c=1}^{C} \langle S^c, G^c \rangle}{\sum_{c=1}^{C} \left( \| S^c \|_1 + \| G^c \|_1 \right)},
\end{equation}
where \( \langle \cdot, \cdot \rangle \) denotes the dot product and \( \| \cdot \|_1 \) the \( \ell_1 \)-norm (sum of elements).

\noindent Focal Loss applies a balancing factor \( \alpha_c \) and focusing parameter \( \gamma \) to emphasize hard examples:
\begin{equation}
\mathcal{L}_{\text{Focal}} = - \sum_{c=1}^{C} \alpha_c \left(1 - S^c \right)^\gamma G^c \log(S^c),
\end{equation}

\noindent The total loss is computed as:
\begin{equation}
\mathcal{L}_{\text{DiceFocal}} = \mathcal{L}_{\text{Dice}} + \mathcal{L}_{\text{Focal}}.
\end{equation}

\section{Experiments}
\label{sec:experiments}
Our objective is to confirm the efficacy of the newly proposed GeoSAM in enhancing segmentation performance across various metrics. This will be accomplished by conducting a comprehensive set of experiments designed to answer critical research inquiries. \textbf{\textit{Q1}}: Does GeoSAM outperform the current state-of-the-art (SOTA) methods in terms of performance in mobility infrastructure segmentation? \textbf{\textit{Q2}}: Can GeoSAM demonstrate superior generalizability by performing effectively on previously unseen datasets? \textbf{\textit{Q3}}: Is automated prompt generation necessary in the case of mobility infrastructure segmentation?

\begin{table}[ht]
    \centering
\caption{Details of the datasets used in this study. $D_{\text{train}}$ and $D_{\text{test}}$ are collected from Washington, D.C., while $D_{\text{gen}}$ is used to evaluate generalization performance on Cambridge, MA.}

\label{tab:datasets_descrption}
\small
\resizebox{\columnwidth}{!}{
    \begin{tabular}{|l|c|c|c|c|c|} \hline 
        \multirow{2}{*}{Dataset} & \multirow{2}{*}{Region} & Geographical & Base Image & \#Base & \#Stitched \\  
        & & Coordinates & Size & Images & Images \\\hline \hline 
        \multirow{2}{*}{$D_{\text{train}}$} & \multirow{2}{*}{Washington DC} & 38.905788, -77.045019 & \multirow{2}{*}{(512,\ 512)} & \multirow{2}{*}{2240} & \multirow{2}{*}{560} \\ 
        & & 38.90968, -77.019694 & & & \\ \hline
        \multirow{2}{*}{$D_{\text{test}}$} & \multirow{2}{*}{Washington DC} & 38.8968333, -77.0074118 & \multirow{2}{*}{(512,\ 512)} & \multirow{2}{*}{1184} & \multirow{2}{*}{296} \\ 
        & & 38.906958, -76.988948 & & & \\ \hline
        \multirow{2}{*}{$D_{\text{gen}}$} & \multirow{2}{*}{Cambridge} & 42.360067, -71.144373 & \multirow{2}{*}{(256,\ 256)} & \multirow{2}{*}{38080} & \multirow{2}{*}{2380} \\ 
        & & 42.395258, -71.051704 & & & \\ \hline
    \end{tabular}
}
\vspace{-23pt}
\end{table}

\subsection{Datasets}
\label{datasets}
We define the training dataset as $D_{\text{train}} = \{I_{\text{train}}, G_{\text{train}}\}$, where $I_{\text{train}}$ and $G_{\text{train}}$ represent $n$ images and corresponding segmentation ground truths (masks for roads and pedestrian infrastructure). Similarly, the test dataset is denoted as $D_{\text{test}} = \{I_{\text{test}}, G_{\text{test}}\}$, and the generalization dataset as $D_{\text{gen}} = \{I_{\text{gen}}, G_{\text{gen}}\}$.

These datasets are constructed from high-resolution orthorectified aerial images and publicly available GIS data~\citep{Washington_GIS,Cambridge_GIS}, following the methodology in~\citep{hosseini2023mapping}. Orthorectified tiles are aerial images that have been geometrically corrected to ensure uniform scale and true top-down perspective, enabling accurate spatial measurements. These tiles~\citep{GeologicalSurvey} are downloaded using geographical bounding boxes and appropriate zoom levels (e.g., zoom level 0 spans the entire world). The GIS data, provided by respective local government authorities, contains accurate coordinate information on urban infrastructure such as roads and sidewalks, enabling reliable mask generation for the two infrastructure classes. We additionally perform manual inspection and correction on the generated masks to fix any potential inconsistencies or missing annotations. GeoSAM is trained and tested in separate regions of Washington, D.C., with an additional test conducted in Cambridge, MA, to evaluate generalization. We first download the base images at their native resolutions (Table~\ref{tab:datasets_descrption}) and stitch adjacent base image tiles within each region to form input images of size \(1024 \times 1024\), using zoom level 20. This resolution corresponds to high-detail aerial imagery, capturing fine-grained urban structures like lanes and sidewalks, consistent with standard geographical mapping scales. Figure~\ref{fig:dataset_example} provides a couple of examples, and the dataset preparation is explained further in Appendix A.3.

\begin{figure}[t]
\centering
\vspace{-22pt}
\includegraphics[width=\linewidth]{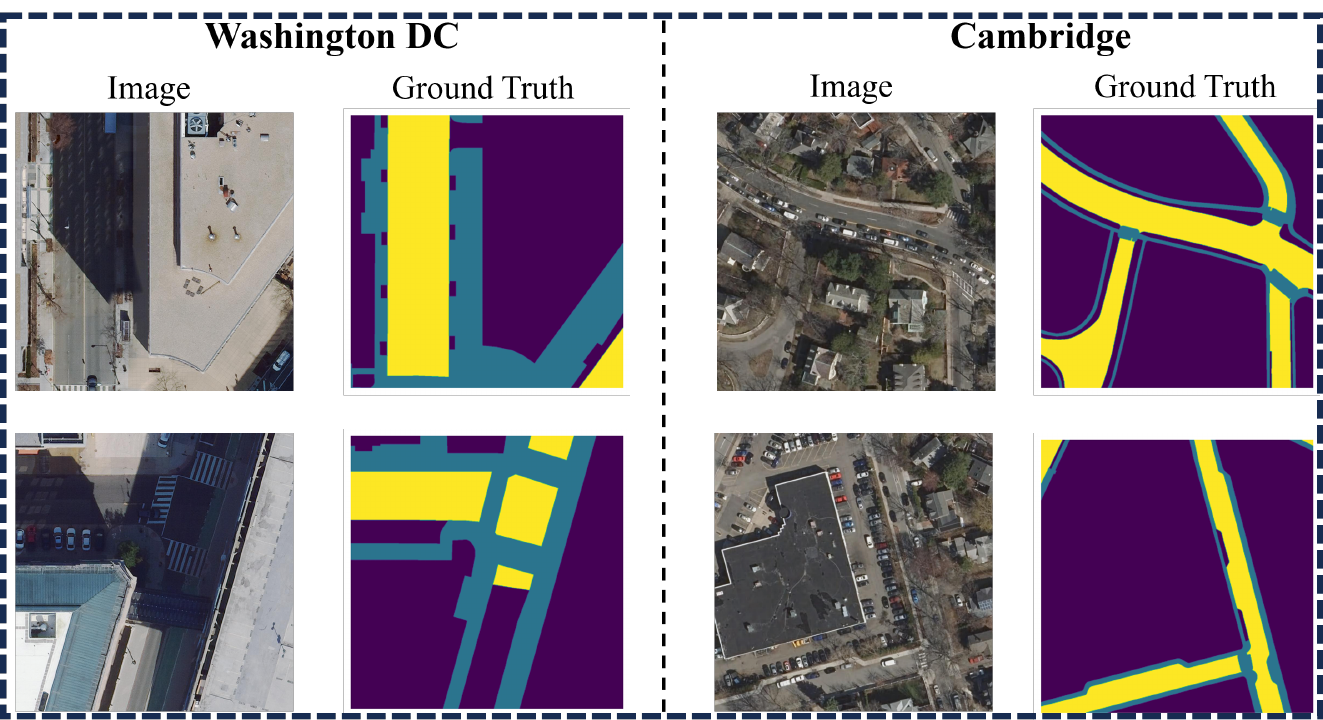}
\vspace{-10pt}
\caption{Randomly picked examples of Washington, D.C. and Cambridge from the datasets (yellow = road, blue = pedestrian infrastructure).}
\label{fig:dataset_example}
\end{figure}

\begin{figure}[t]
\centering
\vspace{-20pt}
\includegraphics[width=\linewidth]{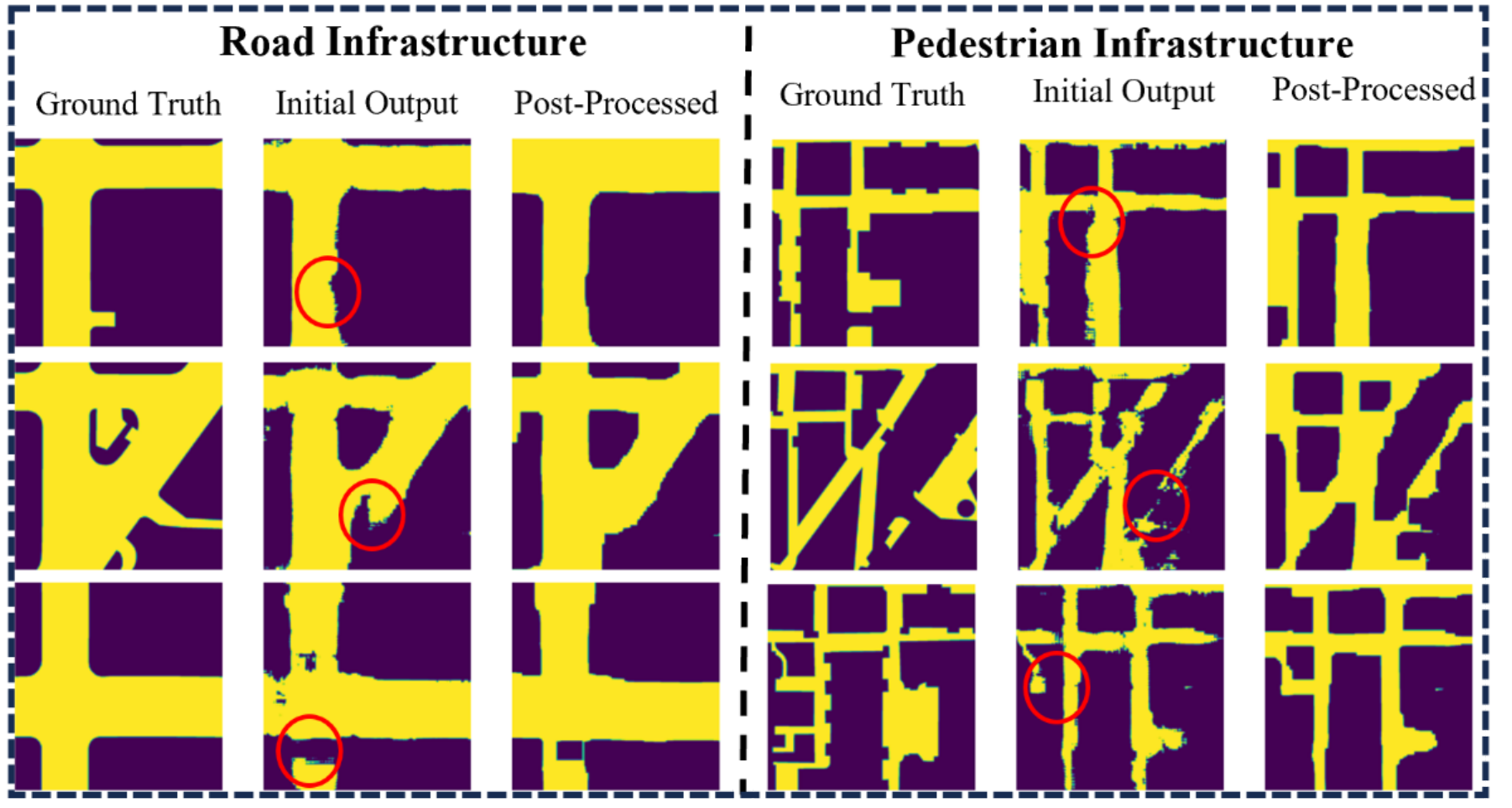}
\vspace{-20pt}
\caption{Postprocessing on two classes, with each row showing a randomly selected test image for (a) road infrastructure and (b) pedestrian infrastructure segmentation. \textcolor{red}{Circles} indicate cosmetic improvements.}

\label{fig:postprocesing}
\end{figure}

\subsection{Implementation Details}


\begin{table*}[t]
\centering
\caption{GeoSAM evaluation results against benchmark models (``Ped.'' for Pedestrian, ``Infras.'' for Infrastructure).  Washington, D.C. used for Testing, and Cambridge, MA used for evaluating generalizability. The best results are in \textbf{bold}, and the second-best results are in \underline{underlined}.}
\small
\resizebox{1.8\columnwidth}{!}{
\begin{tabular}{|l|c|c|c|c|c|c|c|c|c|c|c|c|}
\hline
        \multirow{4}{*}{Method}&\multicolumn{6}{|c|}{$D_{\text{test}}$ (Washington, D.C.)}&\multicolumn{6}{c|}{$D_{\text{gen}}$ (Cambridge, MA)}\\
        \cline{2-13}
        &\multicolumn{2}{c|}{IoU}&\multirow{2}{*}{mIoU}&\multicolumn{2}{c|}{AP}&\multirow{2}{*}{mAP}&\multicolumn{2}{c|}{IoU}&\multirow{2}{*}{mIoU}&\multicolumn{2}{c|}{AP}&\multirow{2}{*}{mAP}\\
        \cline{2-3} \cline{5-6} \cline{8-9} \cline{11-12}
         & Road & Ped.&  & Road & Ped.& & Road & Ped.& & Road & Ped.& \\
        &Infras.&Infras.&&Infras.&Infras.&&Infras.&Infras.&&Infras.&Infras.&\\
        \hline
        \hline
        UNet~\citep{ronneberger2015unet} & 0.45 & 0.17& 0.31&0.44&0.22&0.33&0.24&0.12&0.18&0.11&0.05&0.08\\
        nnU-Net~\citep{isensee2021nnu} & 0.65 & \underline{0.32} & \underline{0.49}& 0.56& 0.32&\underline{0.45} & 0.25& \underline{0.15}&\underline{0.21} &0.05 &  \underline{0.09}& 0.07\\
        UNet++~\citep{zhou2018unet++} & 0.61 & 0.30& 0.45&0.54&\underline{0.34}&0.43&0.24&0.08&0.16&\underline{0.12}&0.06&\underline{0.09}\\
        DeepLabv3+~\citep{chen2018encoder} &  0.47 & 0.18 & 0.32 & 0.46 & 0.22 & 0.34 & 0.10 & 0.06 & 0.08 & 0.05 & 0.04 & 0.04\\
        HRNet~\citep{sun2019deep} &  0.50 & 0.19 & 0.34 & 0.49 & 0.23 & 0.36 & 0.13 & 0.08 & 0.10 & 0.08 & 0.06 & 0.06\\
        \hline
        UNETR~\citep{hatamizadeh2022unetr} & 0.48 & 0.20& 0.34& 0.50&0.27&0.38&\underline{0.27}&0.11&0.18&0.12&0.05&0.08\\
        Swin UNETR~\citep{hatamizadeh2021swin} & 0.63 & 0.26& 0.44& \underline{0.57}&0.29&0.43&0.13&0.09&0.11&0.09&0.04&0.06\\
        SwinUNETR-V2~\citep{he2023swinunetr} & \underline{0.66} & 0.22& 0.43&
        0.54&0.26&0.40&0.15&0.08&0.12&0.09&0.04&0.06\\
        nnFormer~\citep{zhou2021nnformer} & 0.60 & 0.21& 0.41&
        0.52&0.30&0.41&0.16&0.09&0.13&0.11&0.05&0.08\\
        \hline
        Zero-shot SAM~\citep{kirillov2023segment}& 0.30 & 0.18 & 0.24& 0.34& 0.23&0.27 & 0.25& 0.12&0.18 &0.11 &  0.06& 0.08\\
        RSPrompter~\citep{chen2024rsprompter}& 0.46 & 0.20 & 0.33 & 0.49& 0.25 & 0.37 & 0.09& 0.07& 0.08 &0.08 &  0.04& 0.06\\
        UV-SAM~\citep{zhang2024uv}& 0.57 & 0.21 & 0.39 & 0.55& 0.26 &0.40 & 0.11& 0.07&0.09 &0.08 &  0.05& 0.06\\
        GeoSAM (Ours) & \textbf{0.70} & \textbf{0.39} &\textbf{0.54}& \textbf{0.61}&\textbf{0.42}&\textbf{0.51}&\textbf{0.34}&\textbf{0.18}&\textbf{0.26}&\textbf{0.20}&\textbf{0.16}&\textbf{0.18}\\
        \hline
    \end{tabular}
    }
    \vspace{-0.05in}
    \label{tab:geosamresult}
\end{table*}

\noindent \textbf{Experiments Setup} We adopted ViT-H~\citep{dosovitskiy2021vit} as the encoder version of SAM and initialized the model with pre-trained weights from SAM's ViT-H version. Following the original SAM paper settings~\citep{kirillov2023segment}, the choice of optimizer was the AdamW ($\beta_1 = 0.9$, $\beta_2 = 0.999$), with an initial learning rate set at \(10^{-5}\) and weight decay of 0.1, and no data augmentation techniques were applied. Following our experimentation on various values, we chose 0.8 for the balancing factor ($\alpha$) and 2 for the focusing parameter ($\gamma$)  loss function. To have an adaptable learning rate, a cosine annealing learning rate scheduler was employed with a maximum learning rate decaying smoothly to a minimum value (\(10^{-7}\)) over the course of training. A pre-trained nnU-Net~\citep{isensee2021nnu} model (trained on the training dataset) has been selected as the pre-trained traditional model for point prompt generation.  We adopted CLIP's ViT-B version~\citep{radford2021learning} as the text encoder. Finally, for point prompt generation, we selected 2000 foreground and 1000 background points. All the experiments were conducted on an NVIDIA GeForce RTX 4090 GPU with 24 GB of memory and Python 3.10.9. We use a total of 100 epochs to train GeoSAM as well as the other baseline models.


\label{postprocessing}
\noindent \textbf{Postprocessing} We apply postprocessing uniformly to all model outputs, aiming to improve the structural coherence of segmentation maps, with a particular emphasis on ensuring path connectivity over precise pixel-wise correctness. Morphological operations like erosion and dilation address common segmentation issues. Erosion removes isolated regions, ensuring cleaner segmentation of pedestrian paths, while dilation connects disjointed paths to improve route continuity. These operations are performed with a (10$\times$10) filter over a (1024$\times$1024) resolution map and iterated 10 times for effective refinement. Figure \ref{fig:postprocesing} illustrates these techniques, showing improved connectivity and alignment with the ground truth. Comparisons between initial and postprocessed outputs highlight the removal of isolated regions and enhanced path continuity.

\begin{figure*}[t]
\centering
\includegraphics[width=0.75\textwidth]{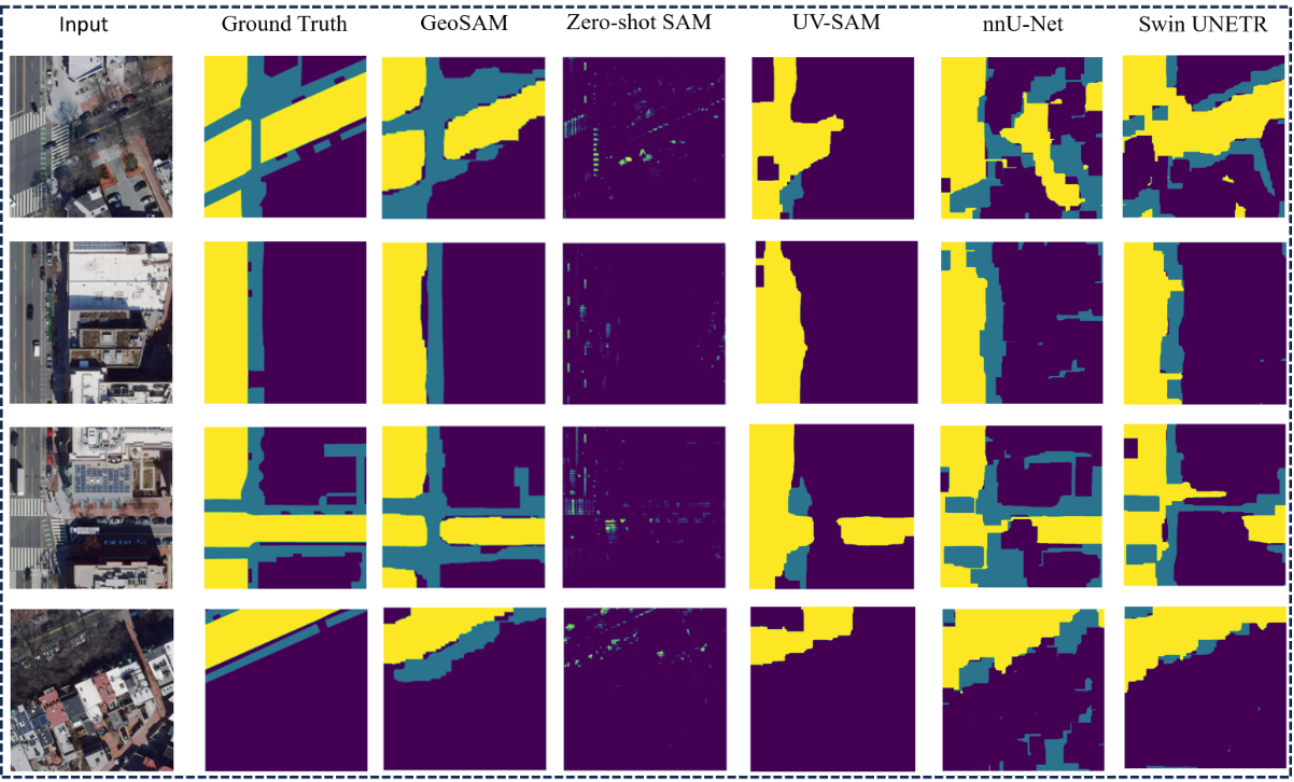}
\vspace{-10pt}
\caption{Comparative qualitative segmentation results: GeoSAM vs. other benchmark models. Different colors (blue=pedestrian infrastructure, yellow=road infrastructure) indicate distinct classes in the multi-class output. Each row displays a randomly selected image from the test dataset.}
\label{fig:output}
\end{figure*}

\noindent \textbf{Benchmark Models} We compare GeoSAM against several popular semantic segmentation models from both CNN- and ViT-based, and SAM-based approaches. All benchmarks follow GeoSAM's training setup, using the same postprocessing and no data augmentation for fair comparison. CNN-based baselines include UNet~\citep{ronneberger2015unet}, nnU-Net~\citep{isensee2021nnu}, UNet++~\citep{zhou2018unet++}, DeepLabv3+~\citep{chen2018encoder}, and HRNet~\citep{sun2019deep}. ViT-based models include UNETR~\citep{hatamizadeh2022unetr}, Swin UNETR~\citep{hatamizadeh2021swin}, SwinUNETR-V2~\citep{he2023swinunetr}, and nnFormer~\citep{zhou2021nnformer}. Additionally, we compare GeoSAM with zero-shot SAM initialized with pre-trained weights (supplemented with point prompts created) and two notable SAM-based geographical segmentation works, such as RSPrompter~\citep{chen2024rsprompter} and UV-SAM~\citep{zhang2024uv}. We train each of the models from scratch using their default settings on $D_{\text{train}}$ described in Section \ref{datasets}. The summary of each of the models can be found in Appendix A.4.
During inference, we evaluate these models on both $D_{\text{test}}$ and $D_{\text{gen}}$ datasets. For each class, we compute the Intersection over Union (IoU) using a fixed threshold to binarize predictions, and the Average Precision (AP) by integrating over all possible thresholds.

\subsection{Results and Discussion}
\label{sec:results}

\noindent\textbf{GeoSAM's Superiority Over Other Methods (\textbf{\textit{Q1}})} Figure \ref{fig:output} presents a qualitative comparison of GeoSAM with zero-shot SAM, UV-SAM, nnU-Net, and Swin UNETR on randomly selected test images. The results highlight the limitations of zero-shot SAM on geographical images outside its training domain, particularly with thin boundary objects, as noted in Section \ref{sec:intro}. In contrast, GeoSAM demonstrates significantly improved segmentation accuracy, closely matching the ground truth. GeoSAM also outperforms UV-SAM, nnU-Net, and Swin UNETR, particularly in handling intricate boundaries by achieving results closest to the ground truth, underscoring its superiority for mobility infrastructure segmentation.

In Table \ref{tab:geosamresult}, we compare GeoSAM's performance with established semantic segmentation models (CNN- and ViT-based) and SAM-based models. On the Washington, D.C. test set, GeoSAM outperforms SOTA models across both classes, surpassing the second-best model, nnU-Net, by 5\% in mIoU and 6\% in mAP. Compared to Zero-shot SAM, GeoSAM achieves a remarkable improvement of 30\% in mIoU and 24\% in mAP, highlighting SAM's limitations with geographical images. GeoSAM significantly outperforms UV-SAM and RSPrompter, the leading SAM-based models in this domain. \textit{These results confirm GeoSAM's effectiveness, even when trained in limited data scenarios.}

\noindent\textbf{Generalization Performance of GeoSAM (\textbf{\textit{Q2}})} GeoSAM's performance on the generalization dataset from Cambridge, MA, reveals a decline across all models due to data shifts between regions. However, GeoSAM consistently outperforms SOTA models, achieving at least double the performance of the second-best model. This highlights the limitations of traditional models, whose generalization is constrained by training data, particularly in visually distinct regions like Cambridge. Foundation models like SAM exhibit better adaptability to diverse scenarios, and GeoSAM, enhanced by automated guided prompts, further improves this adaptability, achieving 5\% and 9\% higher mIoU and mAP than the second-best model. \textit{These results validate GeoSAM's scalability and superior generalization capabilities, leveraging the strengths of a foundation model.}



\noindent\textbf{Necessity of Auto Prompt Generation (\textbf{\textit{Q3}})} Table~\ref{tab:ablation1} shows the effect of varying the number and ratio of point prompts on GeoSAM’s performance.  Since these prompts simulate user input, we identify the optimal configuration generated by our automated system. A 2:1 foreground-to-background ratio with 2000:1000 points performs best, likely due to the large image resolution (\(1024 \times 1024\)), where extensive foreground coverage helps segment large structures like roads and sidewalks while fewer background points reduce ambiguity. We use this as the default in GeoSAM. Even with 150 total points (first row), pedestrian infrastructure performance drops by 8\%, highlighting the model’s sensitivity to the number of prompts. Our chosen 2:1 ratio is further supported by the characteristics of mobility infrastructure: roads and sidewalks are spatially extensive yet sparse, requiring dense sampling across fragmented regions, while the background is semantically redundant. Over‑sampling background reduces the learning signal, but increasing foreground beyond this ratio can lead to over‑segmentation, where disconnected regions may be incorrectly predicted as continuous infrastructure. \textit{These findings underscore the necessity of automated prompt generation, as manually crafting such a large number of prompts is practically infeasible.}

\begin{table}[t]
\centering
\caption{Segmentation performance of GeoSAM using different numbers and foreground-to-background ratios of point prompts.}
\small
\resizebox{\columnwidth}{!}{
    \begin{tabular}{|ccc|cc|}
        \hline
        \multicolumn{3}{|c|}{Point Prompts}& \multicolumn{2}{c|}{IoU}\\
        \hline
        Foreground & Background & \multirow{2}{*}{Ratio} &Road&Pedestrian\\
        Points & Points &  &Infrastructure&Infrastructure\\
        \hline \hline
         100 & 50 & 2:1 & 0.64 & 0.31 \\
         1000 & 500 & 2:1 & 0.65 & 0.33 \\
         2000 & 2000 & 1:1 &0.67& 0.34\\
         \textbf{2000} & \textbf{1000} & \textbf{2:1} & \textbf{0.70}& \textbf{0.39}\\
         2000 & 4000 & 1:2 & 0.66& 0.27\\
        \hline
        
    \end{tabular}
    }

    \vspace{-0.10in}
    \label{tab:ablation1}
\end{table}

\begin{table}[ht]
\centering
\caption{Components Ablation study: examining the effects on performance based on various model components.}
\small
\resizebox{0.8\columnwidth}{!}{
    \begin{tabular}{|ccc|cc|}
        \hline
        \multicolumn{3}{|c|}{Components}& \multicolumn{2}{c|}{IoU}\\
        \hline
        Point&Text&Fine-tuning &Road&Pedestrian\\
        Prompts&Prompts&Decoder &Infras.&Infras.\\
        \hline \hline
        $\checkmark$  & $\checkmark$ & $\checkmark$  & \textbf{0.70}& \textbf{0.39}\\
        $\checkmark$& $\times$& $\checkmark$   & 0.66&0.31 \\
        $\times$ &  $\checkmark$& $\checkmark$  & 0.31&  0.17\\
        $\checkmark$ &  $\times$& $\times$   & 0.24& 0.13\\
        \hline
    \end{tabular}
    }
    
    \label{tab:ablation2}
 
\end{table}

\noindent\textbf{Ablation Study} Table \ref{tab:ablation2} evaluates the impact of key components on GeoSAM, including point prompts, text prompts, and a fine-tuned decoder. Using only point prompts results in an 8\% decrease in pedestrian infrastructure segmentation, emphasizing the critical role of text prompts in providing semantic understanding to resolve ambiguities. However, text prompts alone lead to a 22\% performance drop, demonstrating their insufficiency for nuanced segmentation tasks due to the text encoder's semantic limitations. Instead, text prompts serve as effective secondary prompts, adding context to the decoder. As secondary prompts, they effectively enhance the model's focus by providing additional context to the decoder during segmentation. Removing the fine-tuned decoder further degrades performance, with zero-shot SAM showing a 26\% drop in pedestrian infrastructure segmentation, highlighting the original SAM decoder's inadequacy for geographical images. Fine-tuning adapts the decoder to the unique challenges of this domain.

\begin{table}[ht]
    \centering
    \caption{Backbone Ablation study: performance comparison with different backbones as the traditional pre-trained model to generate automated point prompts for GeoSAM.}
    \resizebox{0.9\columnwidth}{!}{
    \begin{tabular}{|c|c|c|c|c|}
    \hline
         \multirow{2}{*}{Pre-trained}&\multicolumn{2}{c|}{IoU}&\multicolumn{2}{c|}{AP}\\
         \cline{2-5}
         & Road  & Pedestrian& Road  & Pedestrian \\
         Backbone&Infras.&Infras.&Infras.&Infras.\\
         \hline \hline
         UNet~\citep{ronneberger2015unet} & 0.67 & 0.35 & 0.59 & 0.38 \\
         nnU-Net~\citep{isensee2021nnu}& \textbf{0.70} & \textbf{0.39} & 0.61& \textbf{0.42}\\
         Swin UNETR~\citep{hatamizadeh2021swin} & 0.69 & 0.37 & \textbf{0.62} & 0.39\\
         \hline
    \end{tabular}
    }
    \label{tab:more_ablation}
\end{table}

Table~\ref{tab:more_ablation} demonstrates that the choice of backbone used to generate point prompts is not overly critical. We observe only minor performance drops when replacing nnU-Net~\citep{isensee2021nnu} (our default) with UNet~\citep{ronneberger2015unet} or Swin UNETR~\citep{hatamizadeh2021swin}. While the overall framework remains robust, we note that pedestrian infrastructure shows slightly higher sensitivity to backbone changes than roads. Because of its structural variability and weaker visual cues, it can become more reliant on accurate spatial guidance. Nonetheless, as long as the backbone provides reasonably well-positioned foreground and background points, GeoSAM maintains strong performance. Even with different backbones, GeoSAM consistently outperforms all state-of-the-art models listed in Table~\ref{tab:geosamresult}.

\begin{table}[ht]
\centering
\caption{Average inference time of different models on $D_{\text{test}}$ with $1024 \times 1024$ input images. Inference times are reported based on the implementation and settings described in this work; results may vary under different configurations.}

\resizebox{0.8\columnwidth}{!}{
\begin{tabular}{|l|c|c|}
\hline
\multirow{2}{*}{Model} & Tuned &Inference Time \\
&Params. (M) &(sec./image)\\ \hline\hline
nnUNet \citep{isensee2021nnu}  &7.8& 2.01\\
DeepLabV3+ \citep{chen2018encoder}  &5.4 & 2.86\\
Swin UNETR \citep{he2023swinunetr}  & 6.3& 3.44\\
GeoSAM (ours)  &4.2& 3.06\\
\hline
\end{tabular}
}

\vspace{-3mm}
\label{ablation3}
\end{table}

Further, as shown in Table \ref{ablation3}, GeoSAM attains competitive inference speed, being only marginally slower than nnUNet \citep{isensee2021nnu} and DeepLabV3+ \citep{chen2018encoder}, despite leveraging a foundation model-based encoder. This efficiency largely stems from our design choice to pre-compute pseudo labels offline and load them from disk during inference, enabling fast generation of point prompts without additional runtime overhead. Notably, GeoSAM outperforms Swin UNETR \citep{he2023swinunetr} in inference time, underscoring its efficiency among recent SOTA methods. Moreover, GeoSAM requires the fewest tunable parameters during training, highlighting its suitability for resource-constrained settings and real-world deployment.

\section{Conclusion}

GeoSAM adapts SAM for mobility infrastructure segmentation in geographical images, with a strong social impact, particularly for pedestrian safety. It integrates multi-modal prompts (point and text) and fine-tunes SAM’s decoder. Unlike existing methods, our training and end-to-end inference pipeline is transferable across locations and classes using any pre-trained traditional model. The approach is generic, reproducible, and adaptable to various domain-specific segmentation tasks.

\noindent{\textbf{Limitation and Future Work}} We aim to extend the application of GeoSAM to a wider range of geographical regions by incorporating datasets from various other cities, enabling a more comprehensive analysis of its generalizability across diverse urban layouts and visual conditions. In addition, we plan to expand support for additional object types such as stairs, islands/bridges, and potholes, as well as explore its applicability to other imaging modalities.

\begin{ack}
This paper was supported by the U.S. National Science Foundation (NSF) under Award Number 2235225. Any opinions, findings, and conclusions or recommendations expressed in this material are those of the authors and do not necessarily reflect the views of the National Science Foundation.
\end{ack}

\bibliography{references}
\appendix
\section{Appendix: Additional Technical Details and Benchmarks}
\subsection{Experiment Setup}
\textbf{The Intersection over Union (IoU)} score is a ratio of the area of overlap between the predicted and ground truth regions to the area of union. Mathematically, it is defined as:

\[
\text{IoU} = \frac{|\text{Predicted Region} \cap \text{Ground Truth Region}|}{|\text{Predicted Region} \cup \text{Ground Truth Region}|}
\]

where \( |\cdot| \) denotes the cardinality (or area) of the set. The IoU score ranges from 0 to 1, where a value of 0 indicates no overlap and a value of 1 indicates a perfect overlap between the predicted and ground truth regions.

Similarly, \textbf{the Average Precision (AP)} measures the accuracy of the model in classifying each pixel, considering both the precision and the recall across different threshold levels. Precision and recall are defined as:

\[
\text{Precision} = \frac{\text{True Positives (TP)}}{\text{True Positives (TP)} + \text{False Positives (FP)}}
\]

\[
\text{Recall} = \frac{\text{True Positives (TP)}}{\text{True Positives (TP)} + \text{False Negatives (FN)}}
\]

The AP is the area under the Precision-Recall curve, which is computed by varying the decision threshold. The AP can be mathematically expressed as:

\[
\text{AP} = \int_0^1 P(R) \, dR
\]

where \( P(R) \) is the precision as a function of recall. The integral is typically approximated using a finite sum over discrete recall levels:

\[
\text{AP} = \sum_n (R_n - R_{n-1}) P_n
\]

where \( R_n \) and \( R_{n-1} \) are the recall values at thresholds \( n \) and \( n-1 \) respectively, and \( P_n \) is the precision at threshold \( n \).

We utilize both of these because employing both IoU and AP allows for a comprehensive and nuanced evaluation of semantic segmentation models, ensuring that both spatial accuracy and classification performance are adequately assessed.

\subsection{Postprocessing Techniques}

The Equations for erosion and dilation techniques of the postprocessing are described below where $(x,y)$ is a pixel in the image, $B(i, j)$ is the structuring element or the mask to do the operation, $(i, j)$ are the coordinates within the structuring element, $\bigcap$ is the intersection and $\bigcup$ is the union operation:

\begin{equation}
E(x, y) = \bigcap_{(i,j) \in B} I(x+i, y+j),
\end{equation}

\begin{equation}
D(x, y) = \bigcup_{(i,j) \in B} I(x+i, y+j).
\end{equation}

We apply a (10$\times$10) filter across a (1024$\times$1024) resolution segmentation map to perform erosion and dilation on covered regions. These operations are repeated for 10 iterations to refine the map effectively.

\subsection{Dataset Preparation}

To construct our datasets, we begin by selecting target regions using Google Earth to visually identify urban areas of interest. Once the region is selected, we utilize publicly available aerial imagery and GIS data sources, as cited in Section~\ref{datasets} of the main paper. High-resolution orthorectified aerial images are downloaded in the form of base-size tiles using geographic bounding boxes that cover the selected area. These tiles are then stitched together to form composite images of the desired shape and resolution (\(1024 \times 1024\)), suitable for model input. The zoom level is set to 20, which corresponds to high-detail urban-scale resolution, capturing fine-grained structures such as lanes, sidewalks, and curbs. To generate ground-truth masks, we retrieve the corresponding GIS shapefiles for each region. These shapefiles contain precise geospatial coordinates for different infrastructure types—such as roads and pedestrian pathways—defined by local government agencies. We project these vector-based GIS annotations onto the same spatial coordinate system as the stitched aerial images, rasterizing them into pixel-level semantic masks. Each infrastructure class is assigned a distinct label (e.g., road, sidewalk), resulting in class-specific color-coded segmentation maps. Although the data sources are publicly available, the dataset construction process involves extensive preprocessing, including tile stitching, coordinate alignment, and rasterization. We also perform manual verification and correction to ensure annotation quality and resolve missing or noisy labels. The final dataset will be released publicly upon paper acceptance to support reproducibility and further research.

\subsection{Benchmark Models}
\noindent \textbf{UNet}~\citep{ronneberger2015unet}: A classic encoder-decoder architecture with skip connections, where the encoder captures context and the decoder enables precise localization. Widely used in medical and remote sensing segmentation.

\noindent \textbf{nnU-Net}~\citep{isensee2021nnu}: A self-configuring variant of UNet that automatically adapts architectural and training hyperparameters to a given dataset, providing strong out-of-the-box performance without manual tuning.

\noindent \textbf{UNet++}~\citep{zhou2018unet++}: Enhances UNet by introducing nested and dense skip connections, improving gradient flow and feature fusion between encoder and decoder for better multi-scale segmentation.

\noindent \textbf{DeepLabv3+}~\citep{chen2018encoder}: Incorporates Atrous Spatial Pyramid Pooling (ASPP) to capture multi-scale context and uses a decoder module to refine object boundaries, particularly effective for dense semantic segmentation tasks.

\noindent \textbf{HRNet}~\citep{sun2019deep}: Maintains high-resolution representations throughout the network via parallel multi-scale branches and repeated fusion, enabling fine-grained segmentation with detailed spatial precision.

\noindent \textbf{UNETR}~\citep{hatamizadeh2022unetr}: Uses a ViT encoder to model long-range dependencies and integrates features at multiple levels into a CNN-based decoder, enabling accurate segmentation of 3D medical images.

\noindent \textbf{Swin UNETR}~\citep{hatamizadeh2021swin}: Replaces the ViT encoder in UNETR with a Swin Transformer, leveraging hierarchical representations and shifted window attention for improved efficiency and locality.

\noindent \textbf{SwinUNETR-V2}~\citep{he2023swinunetr}: An enhanced version of Swin UNETR with architectural improvements such as multi-resolution fusion and advanced attention modules for better performance and efficiency.

\noindent \textbf{nnFormer}~\citep{zhou2021nnformer}: Combines transformer-based encoding with hierarchical patch embeddings and convolutional upsampling, designed for improved global reasoning and effective 3D segmentation.

\noindent \textbf{RSPrompter}~\citep{chen2024rsprompter} introduces a prompt learning approach for adapting SAM to remote sensing instance segmentation. It extracts multiscale features from intermediate layers of SAM’s frozen image encoder and passes them through a lightweight prompt generator that produces semantic-aware prompt embeddings. These learned prompts are then injected into SAM’s mask decoder to perform category-specific instance segmentation without requiring manual prompts.

\noindent \textbf{UV-SAM}~\citep{zhang2024uv} adapts SAM for urban village segmentation by employing a lightweight specialist segmentation model (e.g., SegFormer) to generate four types of category-specific prompts: mask, box, image embedding from the specialist, and image embedding from SAM. A prompt mixer module fuses these into a unified embedding, which is used to guide SAM’s frozen mask decoder for fine-grained segmentation. This generalist-specialist framework enhances boundary precision and enables SAM to perform well on weakly structured objects like urban villages.

\begin{algorithm}
\scriptsize
\caption{GeoSAM: Training}
\begin{algorithmic}[1]
\Procedure{Train}{$loader$, $sam$, $utils$, $text\_embed$, $auto\_point\_generation$}
    \State $sam.train()$
    \State $optimizer \gets utils['optimizer']$
    \State $loss\_func \gets utils['loss']$
    \For{$iter \in range(epochs)$}
        \For{$image, pseudo\_labels, gt \in [loader]$}
            \State $image\_embed \gets sam.image\_encoder(image)$
            \State $points \gets auto\_point\_generation(pseudo\_labels)$
            \State $points\_embed \gets sam.prompt\_encoder(points)$
            \State $prompts\_embed \gets concatenate(points\_embed, text\_embed)$
            \State $logits \gets sam.decoder(image\_embed, prompts\_embed)$
            \State $loss \gets loss\_func(logits, gt)$
            \State $optimizer.zero\_grad()$
            \State $loss.backward()$
            \State $optimizer.step()$
        \EndFor
    \EndFor
\EndProcedure
\end{algorithmic}
\label{alg}
\end{algorithm}

\section{The Algorithm}

Algorithm~\ref{alg} outlines the training procedure of GeoSAM. The overall pipeline extends the Segment Anything Model (SAM) by incorporating task-specific prompts—both spatial (point-based) and semantic (text-based)—and training only the decoder while keeping the rest of SAM frozen, following a parameter-efficient fine-tuning strategy. Each iteration of the training loop begins by passing the input image through SAM’s frozen image encoder to generate image embeddings. The corresponding pseudo-labels, obtained from a traditional segmentation model, are used to automatically generate a set of structured point prompts via a pre-defined heuristic. These point prompts are then encoded using SAM’s prompt encoder to obtain spatial embeddings. These are concatenated with fixed text embeddings for each target class, forming a joint prompt embedding. The SAM decoder receives the image embeddings and joint prompt embeddings to predict class-specific segmentation logits. The prediction is supervised using the ground-truth segmentation mask, and the combined Dice Focal loss is computed. The decoder weights are then updated via backpropagation. This algorithm enables the GeoSAM framework to be fine-tuned efficiently across different datasets and geographies, without modifying the frozen components of the base foundation model. The modularity of the training procedure also allows for flexible integration of different text embeddings, point generation strategies, or segmentation backbones used for generating pseudo-labels.



\end{document}